\documentclass{article} 
\usepackage[final]{colm2026_conference}

\usepackage{microtype}
\usepackage{graphicx}
\usepackage{subcaption}
\usepackage{booktabs}
\usepackage{hyperref}
\usepackage{url}

\usepackage{lineno}

\definecolor{darkblue}{rgb}{0, 0, 0.5}
\hypersetup{colorlinks=true, citecolor=darkblue, linkcolor=darkblue, urlcolor=darkblue}

\usepackage{tcolorbox}
\tcbuselibrary{skins}
\definecolor{blue-back}{RGB}{235,243,255}
\definecolor{blue-title}{RGB}{50,100,200}
\definecolor{orange-back}{RGB}{255,245,235}
\definecolor{orange-title}{RGB}{200,120,50}
\definecolor{green-back}{RGB}{235,250,240}
\definecolor{green-title}{RGB}{50,150,80}

\usepackage{amsmath}
\usepackage{amssymb}
\usepackage{mathtools}
\usepackage{amsthm}
\usepackage{multirow}
\usepackage{arydshln}

\usepackage[capitalize,noabbrev]{cleveref}

\theoremstyle{plain}

\theoremstyle{definition}

\theoremstyle{remark}

\usepackage[textsize=tiny]{todonotes}

\title{Hallucination Self-Play: Bootstrapping Reinforced Detector via Evolved Generator}

\author{
\normalfont
\parbox{\textwidth}{\raggedright
\rule{0pt}{24pt}%
{\bfseries Shiping Yang$^{1,2}$\thanks{Work done during an internship at Microsoft} \quad
Shining Liang$^{2}$ \quad
Weihao Liu$^{3}$ \quad
Wenbiao Ding$^{2}$}\\[2pt]
{\bfseries Linjun Shou$^{2}$ \quad
Lu Cheng$^{3}$ \quad
Angel X.\ Chang$^{1}$}\\[6pt]
$^{1}$Simon Fraser University \quad $^{2}$Microsoft \quad $^{3}$University of Illinois at Chicago}
}

\begin{document}

\ifcolmsubmission
\linenumbers
\fi

\maketitle

\begin{abstract}
  Identifying faithfulness hallucinations in LLM-generated outputs remains challenging due to the scarcity of high-quality annotated data. Recent work relies on advanced LLMs to synthesize training data, including rationales, labels, and hallucinated claims. However, these methods treat the generator as a static component, limiting iterative improvement of the detector.
  To address this limitation, we introduce Hallucination Self-Play (HSP), a novel framework that enables the detector to bootstrap with an evolved generator.
  HSP involves two roles initialized from the same base model, a detector that assesses the faithfulness of model outputs, and a generator that produces increasingly hard-to-detect hallucinated responses.
  Specifically, the detector is first fine-tuned on human-labeled data and then employed as a reward model to train the generator via reinforcement learning from AI feedback (RLAIF). In turn, the evolved generator synthesizes hallucination data to further optimize the detector through rule-based reinforcement learning.
  Experiments on RAGTruth and LLM-AggreFact across three model families demonstrate that the proposed framework can progressively enhance a small LLM to match or even outperform advanced LLMs without external supervision. Our code is available at {\footnotesize\url{https://github.com/maybenotime/Hallucination_Self-Play}}.
\end{abstract}

\section{Introduction}
\label{sec:intro}
Despite the remarkable capabilities of large language models (LLMs) across diverse domains~\citep{achiam2023gpt,li2025generation,zhang2025find}, they remain prone to hallucinate when handling long-tail knowledge or outdated information~\citep{kandpal2023large,vu2024freshllms}. Retrieval-augmented generation (RAG) has emerged as an effective paradigm to improve the factuality of model responses by grounding them in retrieved documents~\citep{ragsurvey,yang2025quantifying}. However, even with RAG, LLMs still suffer from faithfulness hallucinations, i.e., generating claims that are contradictory to or unsupported by the provided context~\citep{knowledge-conflicts,halusurvey}. Therefore, detecting such hallucinations is critical for providing trustworthy LLM services.


Prior work leverages advanced LLMs to determine whether a model response contains hallucinations~\citep{yang2023new,manakul2023selfcheckgpt,dhuliawala2024chain}. While these methods achieve impressive performance, they are impractical for real-world application due to high inference cost and latency.
This has motivated the development of lightweight and specialized detectors for efficient hallucination detection~\citep{zha-etal-2023-alignscore,ravi2024lynx,seo2025verifying}.
However, the high cost and scarcity of human annotation limit further performance scaling of detectors. To address this, recent studies directly synthesize hallucinated claims using tailored generation pipelines~\citep{cao2023autohall,tang2024minicheck,tan2024large,lei2025factcg}.
A key limitation of such approaches is that the hallucination generators are typically static, lacking adaptivity to the evolving capabilities of the detector.
As a result, detectors quickly reach a performance plateau, as the synthetic hallucinations become too easy to provide effective training signals for further improvement.

To overcome this limitation, we propose \textbf{Hallucination Self-Play (HSP)}, a closed-loop interaction between two roles: a \textit{generator} and a \textit{detector}, both initialized from the same base model. 
The generator is optimized to produce diverse and challenging hallucinations based on the detector’s feedback, while the detector is trained via RLVR on the resulting synthetic data. This interaction enables both roles to co-evolve without external supervision.

While self-play has achieved great success in large language models for easy-to-verify tasks such as code generation and mathematical reasoning~\citep{cheng2024self,zhao2025absolutezero,chen2025spc,liang2026sws}, its application to hallucination detection remains underexplored due to the fundamental challenge of verifying the validity and correctness of generated hallucinations.
Without a reliable verification mechanism, the training process becomes vulnerable to reward hacking, where the generator may exploit superficial shortcuts to obtain rewards rather than producing genuinely challenging hallucinations.
To this end, we utilize the ground truth labels from QA dataset as a proxy for verification and introduce additional safeguards.

We evaluate HSP on the RAGTruth benchmark under two settings.
In the \textit{Detector w/o CoT} setting, we show that HSP serves as a plug-and-play method that further improves a fine-tuned detector, even with only a single round.
In the more challenging \textit{Detector w/ CoT} setting, we demonstrate that self-play can enable a small model to achieve performance comparable to advanced LLMs in a fully self-bootstrapping manner, without external rationale supervision. Additional evaluation on LLM-AggreFact demonstrates the generalization of HSP across benchmarks. 

Our contribution can be summarized as three folds:
(1) We extend self-play paradigm to hallucination detection, overcoming the limitation of static generators.
(2) We introduce additional verification mechanisms to mitigate reward hacking. Further ablation studies confirm their effectiveness in suppressing such behavior, which is critical for maintaining synthetic data quality and stable detector training.
(3) Experiments demonstrate that our framework achieves strong performance and enable fully self-bootstrapped learning.

\section{Related Work}
\label{sec:related_work}
\subsection{Hallucination Detection.}
Existing approaches to faithfulness hallucination detection mainly follow two directions.
The first prompts a strong LLM to judge whether a generated response is grounded in its source context~\citep{dhuliawala2024chain,jacovi2025facts}.
Although effective, these methods rely on advanced LLMs, making them prohibitively expensive to deploy.
The second line of work instead trains lightweight and deployable detectors~\citep{ravi2024lynx,belyi-etal-2025-luna,kovacs2025lettucedetect}. Given the scarcity of human-annotated data, recent studies have turned to synthetic hallucination generation to obtain more capable detection models~\citep{utama2022falsesum,cao2023autohall}. 
MiniCheck~\citep{tang2024minicheck} synthesizes training data using advanced LLMs, while FactCG~\citep{lei2025factcg} further increases data complexity by constructing multi-hop claims from knowledge graphs. FaithLens~\citep{si2026faithlens} distills explanations from a
reasoning model and optimizes prediction accuracy and explanation quality jointly with
reinforcement learning.

However, existing methods rely on static generators, whose fixed hallucination patterns gradually become easy for the detector and limit further improvement.
To address this, we propose a hallucination self-play framework that bootstraps the detector with an evolving generator.

\subsection{Self-Play.}
Self-play is a paradigm in reinforcement learning, where an agent improves by interacting with copies of itself or co-evolving counterparts~\citep{schmidhuber2013powerplay,schaul2024boundless}. This paradigm became popular following the significant advances of AlphaGo~\citep{silver2017mastering} and AlphaZero~\cite{silver2017zero}, which demonstrated that self-play alone can yield superhuman performance in complex decision-making tasks without human supervision.

More recently, self-play has been actively explored in the context of large language models. Language Self-Play~\citep{kuba2025language} propose a Challenger-Solver framework in which a single model alternately poses tasks and solves them, improving without any human-labeled data. Absolute Zero~\citep{zhao2025absolutezero} and R-Zero~\citep{huang2026r} extend this
paradigm to reasoning, where the model generates its own problems and relies on automatic
verification to bootstrap itself from zero data.
SPC~\citep{chen2025spc} formulates self-play as an adversarial game~\citep{goodfellow2020gan} between a generator and a critic, sharpening the critic’s capabilities to detect misleading reasoning steps.

Despite these advances, most existing self-play work targets verifiable domains such as code and math, where automatic verifiers provide reliable rewards.
In contrast, our work applies self-play to hallucination detection, a domain that lacks such off-the-shelf verifiers, and introduce a Hallucination Self-Play framework that enables the detector to continuously self-improve.

\section{Methodology}
\label{sec:method}
\begin{figure}[t]
    \centering
    \includegraphics[width=\linewidth]{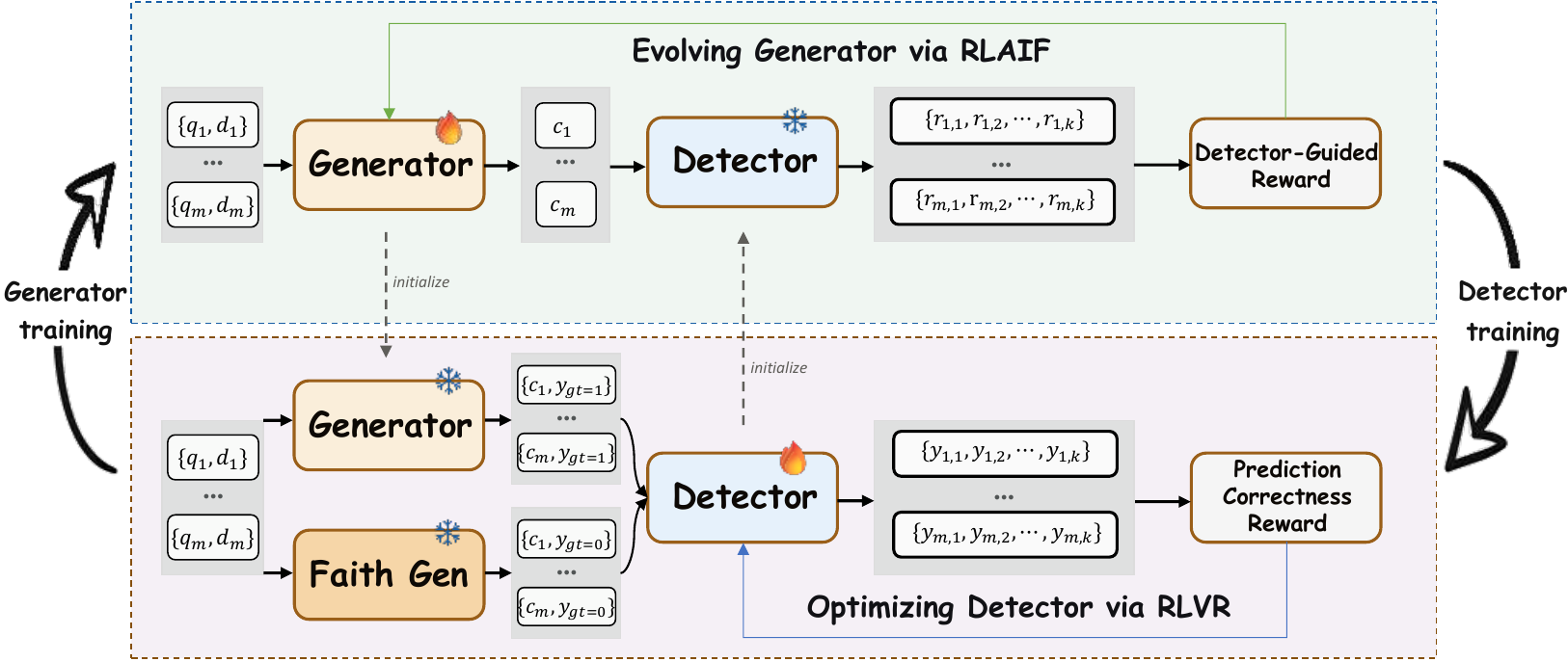}
    \caption{Overview of the HSP framework. The generator is evolved via RLAIF, with the frozen detector providing reward signals. Components for mitigating reward hacking are omitted for clarity (Top). The detector is optimized via RLVR on synthetic data produced by the frozen generators. \textit{Faith Gen} denotes the base model $\mathcal{M}$ prompted to generate faithful responses (Bottom). The two roles interact in a closed loop, with each role alternately frozen while the other is trained.}
    \label{fig:hsp_overview}
\end{figure}

In this section, we propose \textbf{H}allucination \textbf{S}elf-\textbf{P}lay (HSP), a framework that bootstraps the detector by evolving a generator to automatically synthesize hard-to-detect hallucinated responses (Figure~\ref{fig:hsp_overview}).
We first formulate the tasks of the detector and the generator, and detail how these two roles are initialized from the same base model (\S~\ref{sec:two_roles}). Next, we introduce the training algorithms and the corresponding reward design for both roles (\S~\ref{sec:rl_training}). Finally, we describe the closed-loop self-play procedure, illustrating the iterative interaction between the detector and the generator (\S~\ref{sec:self_play_loop}).

\subsection{Two Roles: Detector and Generator}
\label{sec:two_roles}
The HSP framework consists of two interacting roles: a detector and a generator.
Both roles are instantiated from the same base model but are specialized for different inputs, outputs, and training objectives.

\subsubsection{Task Formulation}

\paragraph{Hallucination Detection.}
Given a grounding document $doc$ and an LLM-generated claim $c$, the detector model $D$ is responsible for determining whether the claim is faithful or hallucinated. We consider a claim to be faithful if it is fully supported by the grounding document. Conversely, a claim is regarded as hallucinated if it contradicts or cannot be verified using the provided context.
While previous work typically formulates hallucination detection as a binary classification task~\citep{yang2023new,tang2024minicheck,seo2025verifying}, many real-world applications require locating the specific hallucinated spans. To this end, following ~\citet{niu2024ragtruth}, we train the model to generate a structured output sequence $z$, which represents hallucinated spans in a JSON format. Formally, the detector $D$ models the conditional probability:

\begin{equation}
P_D(y, z \mid doc, c),
\label{eq:detector_prob}
\end{equation}

where y is a binary label derived from $z$, with $y = 0$ if $z$ is empty (i.e., no hallucinated spans are identified), and $y = 1$ otherwise.
Recent work suggests that learned reasoning process can facilitate hallucination span detection~\citep{su2025learntoreason}.
Therefore, we further extend the detector to a variant that not only predicts hallucinated spans, but also generates a rationale $cot$ (i.e., Chain-of-thought) to justify its prediction. This variant can be formulated as:

\begin{equation}
P_D(y, z, cot \mid doc, c).
\end{equation}

\paragraph{Hallucination Generation.}
Given a query $q$ and its corresponding grounding document $doc$, the generator model $G$ is designed to intentionally synthesize hallucinated claims. Such synthetic claims are constructed to appear plausible yet contain inconsistencies with the given document. Formally,

\begin{equation}
P_G(c \mid q, doc, y = 1),
\label{eq:generator_prob}
\end{equation}

where $y=1$ represents that the generator is conditioned to generate hallucination.

\subsubsection{Roles Initialization}
This stage aims to equip the model with basic capabilities for hallucination detection and hallucination generation, serving as an initialization for the subsequent self-play reinforcement learning stage.

\paragraph{Initialize Detector.}
We cold-start the detector $D$ by training the base model $\mathcal{M}$ via supervised fine-tuning (SFT) on a small dataset $\mathcal{D}_{\text{sft}}$.
We consider two variants of the detector and describe how their SFT data is obtained.
For the detector variant without chain-of-thought, $\mathcal{D}_{\text{sft}}$ is sourced from the human-annotated RAGTruth dataset~\citep{niu2024ragtruth}. Each training instance consists of a grounding document $doc$, a generated claim $c$, and a structured label $z$ with annotated hallucinated spans. 

For the detector variant with chain-of-thought, existing training datasets provide only prediction labels but lack rationale annotations. A common approach to bridge this gap is to leverage advanced LLMs to synthesize data with rationales for distillation~\citep{song2024raghat}. In contrast, we investigate whether a small LLM can self-bootstrap its reasoning capabilities for hallucination span detection in the absence of annotated rationales data.
Therefore, we employ a rejection sampling strategy to mine high-quality reasoning paths from the model itself, using span-level annotations as the filtering signal.
Specifically, for each $(doc, c)$ pair, we sample multiple candidate rationales from the base model $\mathcal{M}$ by instructing it to generate reasoning traces alongside hallucinated spans. Candidates that fail to identify all labeled hallucination spans are rejected, and only those whose predicted spans fully cover the ground truth are retained. The resulting samples are treated as pseudo-labeled rationale data and aggregated to form $\mathcal{D}_{\text{sft}}$. 
Given the constructed SFT dataset, we finetune the detector to maximize the conditional likelihood of the target output $o$:

\begin{equation}
\mathcal{L}_{\text{sft}}(\theta_D)
=
-\mathbb{E}_{(doc,c,o)\sim \mathcal{D}_{\text{sft}}}
\left[
\log P_D(o \mid doc, c)
\right],
\label{eq:sft_loss}
\end{equation}

where $o=z$ for the detector without CoT, and $o=(z,cot)$ for the detector with CoT.

\paragraph{Initialize Generator.}
We initialize the generator $G$ from the same base model $\mathcal{M}$ using a prompt-based approach.
Previous work often relies on predefined templates or heuristic rules to synthesize hallucinations, while we do not impose any fixed hallucination strategy in the prompt template. This initialization encourages diverse hallucination patterns to emerge dynamically through interaction with the detector. The prompt used for the generator is shown in Figure~\ref{fig:generator_template}.




        



\subsection{Reinforcement Learning Training}
\label{sec:rl_training}
While SFT provides a strong initialization for the detector, it is ill-suited for subsequent self-play training.
For the detector, annotated hallucination spans and reasoning paths are difficult to obtain from the generator.
Moreover, SFT tends to encourage mode covering, which can harm generalization when trained on "adversarial" samples produced by the generator, whereas reinforcement learning favors mode-seeking behavior~\citep{chen2025retaining}.
For the generator, the objective of synthesizing hard-to-detect hallucinations cannot be explicitly supervised, as the generator only receives scalar feedback from the detector.
Therefore, we adopt reinforcement learning for training both roles in hallucination self-play.

\subsubsection{Policy Optimization Algorithm}
Policy optimization methods for reinforcement learning of LLMs~\citep{yang2026modularized}, such as PPO~\citep{schulman2017ppo} and GRPO~\citep{shao2024grpo}, have been well studied. Given the advantage of GRPO, including the removal of the critic model, we utilize GRPO to optimize both roles in our HSP framework. Under this unified optimization algorithm, the two roles differ only in their input-output formats and reward functions.

For each training instance $x$, we sample a group of $K$ candidate outputs $\{o_1, \ldots, o_k\}$ from the old policy $\mathcal{M}_{\text{old}}$. Each candidate is then evaluated by a role-specific reward function, producing reward $\{r_1, \ldots, r_k\}$. GRPO estimates an advantage $A_i$ for each output using relative rewards within the group, guiding policy updates according to the following objective:

{\scriptsize
\begin{equation}
\mathcal{L}_{\text{GRPO}}(\mathcal{M}_{\theta})
=
\mathbb{E}_{x,\{o_i\}\sim \mathcal{M}_{\text{old}}}
\!\left[
\frac{1}{K}\sum_{i=1}^{K} \min\!\left(
w_i A_i,\;
\mathrm{clip}(w_i, 1-\epsilon, 1+\epsilon)\, A_i
\right)
-
\beta\, \mathbb{D}_{\mathrm{KL}}\!\left(
\mathcal{M}_{\theta} \,\|\, \mathcal{M}_{\text{ref}}
\right)
\right],
\label{eq:grpo}
\end{equation}
}

where $w_i = \frac{\mathcal{M}_{\theta}(o_i \mid x)}{\mathcal{M}_{\text{old}}(o_i \mid x)}$. Here, $\mathcal{M}_{\text{ref}}$ denotes the reference policy (i.e., the initialized model), $\mathcal{M}_{\text{old}}$ is the policy before the update, $\epsilon$ controls the clipping range, and $\beta$ is the coefficient for KL regularization.

\subsubsection{Evolving Generator via RLAIF}
Previous work has shown that task difficulty is crucial for effective reinforcement learning~\citep{zeng2025simplerl,yu2025dapo}. 
To enable the continuous improvement of the detector, we evolve the generator to produce hard-to-detect hallucinations using feedback from the detector, i.e., reinforcement learning from AI feedback (RLAIF).

\paragraph{Detector-Guided Reward.}
Inspired by~\citet{zhao2025absolutezero}, we design a detector-guided reward that encourages the generation of hard-but-solvable hallucinations.
Specifically, we rollout the detector $K$ times with a high sampling temperature\footnote{In our experiments, we sample $10$ times with temperature $1.0$.} and estimate the average success rate $\hat{r}_{\text{acc}}$ as a proxy for the learnability of a synthetic hallucination. This reward can be defined as:

\begin{equation}
\label{eq:detector-guided_reward}
r_{\text{detector-guided}} =
\begin{cases}
0, & \text{if } \hat{r}_{\text{acc}} = 0, \\
1 - \hat{r}_{\text{acc}}, & \text{otherwise},
\end{cases}
\end{equation}

where $\hat{r}_{\text{acc}} = \frac{1}{K} \sum_{i=1}^{K} r_{\text{acc}}^{(i)}$, and $r_{\text{acc}}^{(i)}$ is a binary indicator of whether the detector correctly identifies the synthetic hallucination in the $i$-th rollout.

\paragraph{Reward Hacking.}
When the generator is optimized solely with the reward defined in Eq.~\ref{eq:detector-guided_reward}, the training objective is highly vulnerable to reward hacking. Specifically, the generator can easily finds a shortcut strategies that maximize the expectation of reward without actually learning to produce high-quality hallucinations. A common failure mode is that the generator outputs a faithful answer, which the detector consequently fail to identify as hallucinated with high accuracy, thereby yielding a high reward. This hacking behavior introduces incorrect labels into the synthetic data, causing the detector to receive noisy reward and fail to learn robust decision boundaries.

\paragraph{Reward Gating Criteria.}
To address the reward hacking issue, we introduce two reward gating criteria to verify whether the generated response is indeed hallucinated. These criteria are motivated by the definition of faithfulness hallucination. A response that satisfies at least one criterion is considered hallucinated and receives a reward, while a response that fails all criteria receives a penalty.
This design prevents the generator from gaming the reward function by producing non-hallucinated responses to obtain high rewards.

First, a generated claim is gated as eligible for detector-guided reward if it contradicts the provided context.
We use the ground-truth answer from a QA dataset as a simplified indicator to assess whether the generated response conflicts with the given context. Specifically, the response is deemed inconsistent with the grounding document if the correct answer is absent, and thus passes this reward gate. In addition to a basic string-matching check, we implement a model-based method that instructs an LLM to identify whether the generated response contains aliases or variants of the answer. The prompt can be found in Figure~\ref{fig:alias_detection_prompt}.

Second, a generated claim is also gated as eligible for detector-guided reward if it introduces facts that are unsupported by the grounding documents.
To implement this criterion, we employ a named entity recognition (NER) model\footnote{We use the NER model \textit{en\_core\_web\_lg} provided by spaCy.} to extract entities from the query, the grounding documents, and the generated response. If the response introduces entities that do not appear in either the query or the given context, it passes this reward gate as containing unsupported facts. 

\paragraph{Trivial Answer Penalty.}
Even with the above gating criteria, we observe an unexpected form of reward hacking: the generator learns to produce trivial or meaningless response, or to refuse answering. Such responses are neither hallucinated nor correct, thereby bypass the reward gating criteria. 
To address this issue, we introduce a hybrid detection strategy: refusal responses are identified via keyword-based heuristics, while trivial or evasive outputs are detected with a model-based approach\footnote{We use the base model $\mathcal{M}$ for all model-based implementations to avoid introducing any external supervision. The prompt template is provided in Figure~\ref{fig:evasive_detection_prompt}.}. All identified responses are penalized to suppress this behavior.

\paragraph{Overall Reward.}
We integrate the \textit{detector-guided reward}, \textit{reward gating criteria}, and \textit{trivial answer penalty} into an overall reward for optimizing the generator. The generator reward is formulated as:

\begin{equation}
\label{eq:generator_overall_reward}
r_{G} =
\begin{cases}
-1, & \text{if the response is trivial}, \\
\mathbb{I}_{\text{gate}}
\cdot
r_{\text{detector-guided}}, & \text{otherwise},
\end{cases}
\end{equation}

where $\mathbb{I}_{\text{gate}} \in \{0, 1\}$ equals $1$ if the generated response satisfies at least one gating criteria, and $0$ otherwise.
This design imposes hard regularization on the generator, constraining the feasible solution space and restricting non-zero rewards to valid hallucinated responses. As a result, the generator is encouraged to synthesize hard yet learnable hallucinations that provide effective training signals for the detector.

\subsubsection{Optimizing Detector via RLVR}

Reinforcement Learning with Verifiable Rewards (RLVR) has recently achieved great success in enhancing the reasoning capabilities of LLMs~\citep{guo2025deepseekr1}. Recent research further demonstrates its effectiveness in hallucination span detection~\citep{su2025learntoreason}. Hence, we apply RLVR to optimize the detector in our HSP framework.

\paragraph{Prediction Correctness Reward.}

\citet{su2025learntoreason} employ a span-level reward function that computes the F1 score between predicted hallucination spans and ground-truth annotations.
However, in our self-play setting, the generator is typically a small LLM. Such models struggle to produce hallucinated claims along with correct span annotations due to limited capabilities. To ensure the data quality, we formulate the generator to produce hallucinated claims only. Accordingly, we adopt a binary prediction correctness reward instead of a span-level reward to optimize the detector.
Formally, given an input pair $(doc, c)$, the detector derives a binary prediction $\hat{y} \in \{0,1\}$, where $\hat{y}=0$ indicates a faithful claim and $\hat{y}=1$ indicates a hallucinated claim. The detector reward is defined as:

\begin{equation}
r_{\text{detector}} = \mathbb{I}\big[\hat{y} = y_{\text{gt}}\big],
\label{eq:detector_binary_reward}
\end{equation}
where $y_{\text{gt}} \in \{0,1\}$ is the ground-truth label and $\mathbb{I}[\cdot]$ is the indicator function.
Through verifiable reward signals, RLVR encourages the detector to refine its reasoning process and prediction strategy in a self-corrective fashion.

\paragraph{Dataset with Verifiable Labels.}
All hallucination responses synthesized by the generator are automatically labeled as hallucinated. To construct balanced training data with faithful labels, we prompt the base model $\mathcal{M}$ to produce answers that strictly adhere to the provided grounding documents.
Although the base model is not guaranteed to be hallucination-free, we empirically observe that the residual noise becomes acceptable after filtering against the ground-truth answers, resulting in labels that are effective for detector training.
Importantly, we do not directly use the ground-truth answers included in QA datasets for detector training. These answers are typically very short and therefore deviate substantially from the distribution of model-generated response.

\subsection{Hallucination Self-Play Loop}
\label{sec:self_play_loop}

We define hallucination self-play as a closed-loop interaction between a generator and a detector, in which learning signals are derived from their internal interaction rather than external supervision. 
The generator is encouraged to produce hallucinated claims that challenge the target detector, while the detector aims to correctly identify the generated hallucinations.
This interaction yields dual learning signals: detector feedback for updating the generator and data with verifiable label for training the detector. Notably, the two roles do not exchange parameters or gradients; instead, learning is driven solely by generated data and derived rewards, forming a stable closed-loop self-play process.


\paragraph{Information Flow in the Self-Play Loop.}

The self-play training loop is driven by the flow of information between the two roles.
Given a query-document pair, the generator synthesizes hallucinated claims, which are then evaluated by a frozen target detector. The detector’s predictions are processed into a scalar reward signal and returned to the generator as feedback. After the generator is evolved via RLAIF, we freeze the updated generator and use it to produce hallucination candidates for detector training. To improve training efficiency, hallucinations are scored by Eq.~\ref{eq:generator_overall_reward} using the previous detector to mine learnable examples. These selected samples are further combined with non-hallucinated responses generated by the base model, forming a balanced training dataset. The detector then predicts on this synthetic dataset and is optimized via RLVR using prediction correctness rewards.

\paragraph{Dynamic Curriculum via Multi-Round Self-Play.}

Even after a single round of self-play training, hallucinations produced by the updated generator quickly saturate and provide little additional learning signal for the detector. As the detector improves, previously challenging hallucinations become easy, limiting further progress.
In contrast, multi-round self-play induces a dynamic curriculum that continuously adapts the difficulty of synthetic hallucinations to the detector’s evolving capability.
This dynamic adjustment of task difficulty maximizes learning efficiency and enables the generator and detector to iteratively co-evolve under an evolving curriculum, driving continuous improvement without external supervision.


\section{Experiments}
\label{sec:experiments}
\begin{table*}[t]
\centering
\resizebox{\textwidth}{!}{%
\setlength{\tabcolsep}{4pt}
\begin{tabular}{l|ccc|ccc|ccc|ccc}
\hline
\multirow{2}{*}{\textbf{Model}}
& \multicolumn{3}{c|}{\textbf{Question Answering}}
& \multicolumn{3}{c|}{\textbf{Summarization}}
& \multicolumn{3}{c|}{\textbf{Data-to-Text}}
& \multicolumn{3}{c}{\textbf{Avg.}} \\
& Recall & Precision & F1
& Recall & Precision & F1
& Recall & Precision & F1
& Recall & Precision & F1 \\
\hline

\multicolumn{13}{c}{\textbf{State-of-the-Art LLMs}} \\
\hline
GPT-4o w/o CoT & 88.1 & 32.6 & 47.6 & 87.3 & 56.0 & 68.2 & 97.4 & 72.0 & 82.8 & 90.9 & 53.6 & 66.2 \\
GPT-4o w/ CoT & 80.0 & 51.0 & 62.3 & 79.9 & 70.0 & 74.6 & 89.3 & 84.1 & 86.6 & 83.1 & 68.3 & 74.5  \\
DeepSeek-V3.2 w/o CoT & 66.9 & 52.2 & 58.6 & 88.7 & 54.4 & 67.4 & 93.6 & 79.9 & 86.2 & 83.1 & 62.2 & 70.8 \\
DeepSeek-V3.2 w/ CoT & 86.2 & 45.9 & 59.9 & 86.8 & 58.6 & 70.0 & 90.2 & 86.9 & 88.5 & 87.7 & 63.8 & 72.8 \\

\hline
\multicolumn{13}{c}{\textbf{Reasoning Models}} \\
\hline
Qwen3-32B & 88.8 & 37.8 & 53.0 & 86.8 & 57.8 & 69.4 & 89.6 & 83.4 & 86.4 & 88.4 & 59.7 & 69.6  \\
Qwen3-14B & 90.6 & 33.8 & 49.2 & 88.2 & 48.8 & 62.8 & 91.2 & 81.4 & 86.0 & 90.0 & 54.7 & 66.0 \\

\hline
\multicolumn{13}{c}{\textbf{\textit{Finetuned Models (Detector w/o CoT)}}} \\
\hline
RAG-HAT & 73.1 & 76.5 & 74.8 & 59.8 & \textbf{77.7} & 67.6 & 90.3 & \textbf{92.9} & \textbf{91.6} & 74.4  & \textbf{82.4} & 78.0   \\
SFT (Qwen2.5-7B-It)  & 74.4 & 76.3 & 75.3 & 66.2 & 74.2 & 69.9 & 91.7 & 91.2 & 91.5 & 77.4 & 80.6 & 78.9 \\
\textbf{HSP Round 1}$^{\dagger}$ (Qwen)  & \textbf{78.1} & \textbf{76.7} & \textbf{77.4} & \textbf{69.1} & 76.2 & \textbf{72.5} & \textbf{93.1} & 90.0 & 91.5 & \textbf{80.1}  & 81.0 & \textbf{80.5} \\
SFT (Llama-3.1-8B-It)  & 73.8 & 74.7 & 74.2  & 57.8 & 77.1 & 66.1 & 91.4 & 91.7 & 91.5 & 74.3 & 81.2 & 77.3\\ 
\textbf{HSP Round 1}$^{\dagger}$ (Llama) & 76.9 & 72.8 & 74.8 & 62.3 & 77.0 & 68.8 & 92.4 & 90.8 & \textbf{91.6} & 77.2 & 80.2 & 78.4   \\

\hline
\multicolumn{13}{c}{\textbf{\textit{Detector w/ CoT (Self-Bootstrapping without Curated Data)}}} \\
\hline
RSFT & 63.8 & \textbf{72.9} & 68.0 & 58.3 & 75.3 & 65.8 & 55.1 & \textbf{90.6} & 68.5 & 59.1 & \textbf{79.6} & 67.4 \\
RLVR & 67.5 & 72.0 & 69.7 & 60.8 & 78.5 & 68.5 & 61.1 & 88.3 & 72.2 & 63.1 & \textbf{79.6} & 70.1 \\
\textbf{HSP Round 1} & 74.4 & 65.4 & 69.6 & 63.2 & \textbf{80.1} & 70.7 & 72.0 & 85.8 & 78.3 & 69.9  & 77.1 & 72.9 \\
\textbf{HSP Round 2} & 71.9 & 68.5 & 70.1 & \textbf{66.2} & 77.6 & \textbf{71.4} & 74.3 & 86.5 & 79.9 & 70.8  & 77.5 & 73.8 \\
\textbf{HSP Round 3} & \textbf{76.3} & 66.7 & \textbf{71.1}
                   & 63.7 & 78.8 & \textbf{70.5}
                   & \textbf{77.2} & 86.1 & \textbf{81.4}
                   & \textbf{72.4} & 77.2 & \textbf{74.3} \\
\hline
\end{tabular}%
}
\caption{Hallucination detection performance on RAGTruth across three tasks. We report response-level precision, recall, and F1.
For \textit{Detector w/ CoT}, all experiments are conducted using \textbf{Qwen2.5-7B-Instruct} as the base model. \textbf{RSFT} denotes supervised fine-tuning on CoT data mined via rejection sampling, while \textbf{RLVR} further optimizes the RSFT model via RL using hallucination annotations from RAGTruth, without access to labeled rationales. \textbf{HSP} models in this setting are trained iteratively, where each iteration continues training from the model obtained in the last round, forming a multi-round hallucination self-play. For \textit{Finetuned Models}, entries marked with ${\dagger}$ indicate models obtained by applying HSP training on top of the corresponding SFT checkpoint. We highlight the best results within each setting in bold.}
\label{tab:ragtruth-main}
\end{table*}

\begin{table}[t]
\centering
\setlength{\tabcolsep}{3pt}
\resizebox{\columnwidth}{!}{%
\begin{tabular}{lcccccccccccc}
\toprule
\textbf{Model} & \textbf{CNN} & \textbf{XSum} & \textbf{ClaimV} & \textbf{ExpQA} & \textbf{FC-GPT}
& \textbf{LFQA} & \textbf{RAGT} & \textbf{Reveal} & \textbf{T-Med} & \textbf{T-Meet} & \textbf{Wice}
& \textbf{Avg.} \\
\midrule
\multicolumn{13}{l}{\textit{Advanced LLMs}} \\
DeepSeek-V3.2-Think$^\dagger$ & 86.8 & 76.8 & 88.0 & 80.7 & 88.0 & 77.4 & 85.9 & 92.1 & 83.5 & 91.4 & 81.8 & 84.4 \\
GPT-5.2$^\dagger$             & 87.9 & 72.7 & 88.6 & 82.6 & 93.5 & 89.9 & 89.2 & 93.3 & 85.4 & 79.8 & 87.8 & 86.1 \\
\midrule
\multicolumn{13}{l}{\textit{Specialized Detectors}} \\
ClearCheck~\citep{seo2025verifying} & 72.8 & 78.6 & 85.4 & 72.7 & 87.9 & 87.0 & 83.8 & 87.0 & 67.8 & 75.8 & 81.8 & 80.1 \\
FaithLens$^\dagger$  & \textbf{84.9} & \textbf{79.0} & \textbf{89.4} & 79.6 & \textbf{92.4} & 92.1 & 86.8 & 92.2
           & \textbf{85.1} & 87.2 & \textbf{85.6} & \textbf{86.4} \\
\midrule
\multicolumn{13}{l}{\textit{HSP (Qwen3-4B-Instruct-2507)}} \\
SFT          & 66.8 & 74.5 & 85.2 & 80.3 & 86.9 & 88.8 & \textbf{93.5} & 93.3 & 78.2 & 83.8 & 76.4 & 82.5 \\
Round 1  & 70.2 & 75.3 & 86.4 & 81.7 & 85.6 & 93.3 & 92.4 & \textbf{95.6} & 79.7 & \textbf{89.1} & 72.7 & 83.8 \\
Round 2  & 71.4 & 78.5 & 88.6 & \textbf{81.9} & 87.9 & \textbf{94.4} & 88.1 & 93.3 & 80.8 & 83.5 & 79.8 & 84.4 \\
\midrule
\multicolumn{13}{l}{\textit{Ablation: static generator (\S\ref{sec:ablation})}} \\
\quad w/o Evolved Gen. (R1) & 62.3 & 76.7 & 86.4 & 80.5 & 89.0 & 91.0 & 92.4 & 96.7 & 77.4 & 84.4 & 79.8 & 83.3 \\
\quad w/o Evolved Gen. (R2) & 52.0 & 58.1 & 79.9 & 77.4 & 87.9 & 79.8 & 79.4 & 85.0 & 69.4 & 62.6 & 78.1 & 73.6 \\
\bottomrule
\end{tabular}}
\caption{Results on \textbf{LLM-AggreFact}. Among the specialized detectors, FaithLens is the current state of the art; it is built on Llama-3.1-8B-Instruct, with its SFT stage distilled from DeepSeek-V3.2-Think. $\dagger$ indicates scores reported by~\citet{si2026faithlens}. \textit{w/o Evolved Gen.} replaces the RLAIF-evolved generator with the frozen base model while keeping the total amount of RL training fixed. \textbf{Bold} indicates the best result among the specialized detectors and HSP method. Column abbreviations: ClaimV = ClaimVerify, ExpQA = ExpertQA, RAGT = RAGTruth, T-Med/T-Meet = TofuEval-MediaS/MeetB. Note that the RAGT column is a re-labeled RAGTruth subset evaluated with macro-F1 and is \emph{not} comparable to Table~\ref{tab:ragtruth-main}.}
\label{tab:aggrefact}
\end{table}

In this section, we conduct a series of experiments and analyses to demonstrate the effectiveness of our HSP framework.

\subsection{Experiment Setup}
\paragraph{Evaluation.}
We use RAGTruth as evaluation benchmark, which includes three representative tasks in the retrieval-augmented generation setting: \textit{Question Answering} (QA), \textit{Data-to-Text}, and \textit{Summarization}. Each task contains paired source documents, model-generated responses, and human-labeled hallucination spans.
We follow the RAGTruth evaluation protocol and report response-level recall, precision, and f1 scores. For the \textit{Detector w/ CoT} setting, we use the prompt template shown in Figure~\ref{fig:detector_template}.
We additionally evaluate on LLM-AggreFact~\citep{tang2024minicheck} in Section~\ref{sec:aggrefact}, which comprises 11 faithfulness hallucination detection tasks spanning diverse domains.

\paragraph{Implementation Details.}
We use \textbf{Qwen2.5-7B-Instruct} and \textbf{Llama-3.1-8B-Insruct} as the base model in our main results.
For the \textit{Detector w/o CoT} setting, we initialize the detector by fine-tuning the base model on RAGTruth~\citep{niu2024ragtruth}.  
For the \textit{Detector w/ CoT} setting, we instead cold-start the detector using a small set of CoT data mined from \textbf{Qwen2.5-7B-Instruct} using rejection sampling. 
During self-play training, \textit{HotpotQA}~\citep{yang2018hotpotqa} serves as the seed dataset for synthetic data generation.
Additional implementation details and hyperparameters can be found in the appendix~\ref{sec:implementation_details}.

\subsection{Main Results on RAGTruth}

\paragraph{Baselines.}
We compare several baselines, including \textbf{(1) State-of-the-Art LLMs}: the proprietary model GPT-4o~\citep{hurst2024gpt4o} and the advanced open-source model DeepSeek-V3.2~\citep{liu2025deepseekV3};  \textbf{(2) Reasoning Models}: Qwen3-32B and Qwen3-14B, which are optimized for long-cot reasoning~\citep{yang2025qwen3};  \textbf{(3) Supervised Fine-Tuning (SFT)}: the base model trained on the RAGTruth dataset using standard supervised learning; \textbf{(4) RAG-HAT}~\citep{song2024raghat}: a specialized hallucination detection model that outputs binary predictions with detailed explanations, trained on RAGTruth and a curated dataset containing rationales distilled from GPT-4-turbo. 

\paragraph{Detector without CoT.}
We first evaluate HSP in the \textit{Detector w/o CoT} setting by applying one round of self-play on top of SFT checkpoints. As shown in the \textit{Finetuned Models} section of Table~\ref{tab:ragtruth-main}, HSP consistently improves F1 over the corresponding SFT baselines for both \textbf{Qwen2.5-7B-Instruct} and \textbf{Llama-3.1-8B-Instruct} across all three tasks. These results demonstrate that HSP serves as a model-agnostic, plug-and-play post-training strategy. The improvements are primarily driven by higher recall while maintaining comparable precision, suggesting that self-play enables detectors to identify a broader range of hallucinations without substantially increasing false positives.

\paragraph{Detector with CoT.}
We next consider the more challenging \textit{Detector w/ CoT} setting, where neither human-annotated rationales nor external LLM distillation data are available. HSP demonstrates sustained improvements across self-play rounds, with performance increasing from 72.9 (Round 1) to 74.3 (Round 3), reflecting an evolving self-play curriculum.
Although trained solely on QA data, the reasoning capabilities optimized by RLVR show strong cross-task generalization, boosting F1 on the Data-to-Text task from 72.2 to 81.4 (+9.2 points).
Notably, we train a 7B model entirely through self-play without any external rationale supervision, achieving performance comparable to the frontier proprietary model \textbf{GPT-4o w/ CoT} (74.3 vs.\ 74.5). These results show that a compact model can fully self-bootstrap strong hallucination detection capabilities under our framework.

\subsection{Generalization to LLM-AggreFact}
\label{sec:aggrefact}
To assesss the generalization of HSP method, we use the cleaned release of \textbf{LLM-AggreFact}~\citep{tang2024minicheck} from \citet{seo2025verifying} as benchmarks. We follow the evaluation protocol of FaithLens~\citep{si2026faithlens} and report macro-F1 scores. The experiments are conducted with \textbf{Qwen3-4B-Instruct-2507}, following the configuration of \textit{Detector w/ CoT}. To enable efficient study, we reuse the CoT data mined during the RSFT stage of \textbf{Qwen2.5-7B-Instruct} to cold-start the detector, rather than rerunning rejection sampling.

As shown in Table~\ref{tab:aggrefact}, after two rounds of HSP, our method outperforms the SFT baseline on most benchmarks and even surpasses FaithLens on several of them. Although HSP achieves a slightly lower average score than FaithLens (84.4 vs. 86.4), it is fully self-bootstrapping and requires no external rationale supervision or distillation from state-of-the-art models.

\subsection{Ablation Study}
\label{sec:ablation}

We ablate two core design choices of HSP: the evolution of the generator, and the
mechanisms for mitigating reward hacking. Both ablations use \textbf{Qwen3-4B-Instruct-2507} as the backbone model.

\paragraph{Effect of the Evolved Generator.}
A natural question is whether the gains across self-play rounds stem from the co-evolution of the generator and detector, or simply from additional RL updates to the detector.
To isolate this, we construct a \textit{static generator} control: instead of evolving the generator
via RLAIF, we keep the base model frozen and use it to synthesize hallucinations in both rounds,
holding the total amount of RL training fixed. Results are reported in the bottom block of
Table~\ref{tab:aggrefact}.

In the first round, the static generator improves the detector from 82.5 to 83.3 Avg, close to HSP (83.8), showing that a fixed generator can initially provide useful training signal. The second round, however, continued RLVR with the same generator degrades performance to 73.6, while HSP further improves to 84.4. This demonstrates that the multi-round gains come from generator evolution rather than additional RL training alone.

\paragraph{Effect of Reward Hacking Mitigation.}
We next ablate the reward gating criteria and the trivial answer penalty introduced in
§\ref{sec:rl_training} to assess their impact on synthetic data quality. All variants are
evaluated on 500 held-out samples, for which we report the number of positive-reward instances
retained for detector training and their average reward. To further evaluate data quality, we randomly sample 10 examples and manually assess the hallucination rate of the generated responses, defined as the fraction of responses that are genuinely hallucinated.

As shown in Table~\ref{tab:ablation_reward_hacking}, removing the reward gating criteria causes the generator to collapse into producing non-hallucinated responses, achieving a high reward of 1.0 while the actual hallucination rate drops to 0\%. In contrast, the full HSP configuration maintains a high hallucination rate of 90\%, at the cost of lower average reward and fewer retained samples. These results confirm that both components are essential for suppressing reward hacking and ensuring that the synthetic training data remains informative for the detector.


\section{Conclusion}
\label{sec:conclusion}
In this work, we introduce Hallucination Self-Play (HSP), a framework that enables hallucination detectors to self-bootstrap without relying on external supervision. HSP are defined as a closed-loop interaction between a generator and a detector derived from the same base model. The generator is continuously evolved via RLAIF to synthesize challenging yet learnable hallucinations, overcoming the limitations of static generators. The detector is optimized via RLVR on the resulting synthetic data. Experiments on RAGTruth and LLM-AggreFact demonstrate the effectiveness and generalization of HSP.
Overall, HSP shows that self-play offers an effective and scalable paradigm for continuously improving hallucination detectors.


\section*{Ethics Statement}
While our method is designed to improve hallucination detection, the generator component of our HSP framework explicitly learns to produce hard-to-detect hallucinations during training. Such a generator could potentially be misused to generate unfaithful content that is difficult to identify. We emphasize that the generator in our framework is not intended for standalone deployment, but is used exclusively as a controlled component within a hallucination self-play loop.

\bibliography{hallucination_selfplay}

@inproceedings{niu2024ragtruth,
  title={Ragtruth: A hallucination corpus for developing trustworthy retrieval-augmented language models},
  author={Niu, Cheng and Wu, Yuanhao and Zhu, Juno and Xu, Siliang and Shum, Kashun and Zhong, Randy and Song, Juntong and Zhang, Tong},
  booktitle={Proceedings of the 62nd Annual Meeting of the Association for Computational Linguistics (Volume 1: Long Papers)},
  pages={10862--10878},
  year={2024}
}

@inproceedings{yang2023new,
  title={A New Benchmark and Reverse Validation Method for Passage-level Hallucination Detection},
  author={Yang, Shiping and Sun, Renliang and Wan, Xiaojun},
  booktitle={Findings of the Association for Computational Linguistics: EMNLP 2023},
  pages={3898--3908},
  year={2023}
}

@inproceedings{tang2024minicheck,
  title={MiniCheck: Efficient Fact-Checking of LLMs on Grounding Documents},
  author={Tang, Liyan and Laban, Philippe and Durrett, Greg},
  booktitle={Proceedings of the 2024 Conference on Empirical Methods in Natural Language Processing},
  pages={8818--8847},
  year={2024}
}

@inproceedings{lei2025factcg,
  title={FactCG: Enhancing Fact Checkers with Graph-Based Multi-Hop Data},
  author={Lei, Deren and Li, Yaxi and Li, Siyao and Hu, Mengya and Xu, Rui and Archer, Ken and Wang, Mingyu and Ching, Emily and Deng, Alex},
  booktitle={Proceedings of the 2025 Conference of the Nations of the Americas Chapter of the Association for Computational Linguistics: Human Language Technologies (Volume 1: Long Papers)},
  pages={5002--5020},
  year={2025}
}

@article{su2025learntoreason,
  title={Learning to Reason for Hallucination Span Detection},
  author={Su, Hsuan and Hu, Ting-Yao and Koppula, Hema Swetha and Krishna, Kundan and Pouransari, Hadi and Hsieh, Cheng-Yu and Koc, Cem and Cheng, Joseph Yitan and Tuzel, Oncel and Vemulapalli, Raviteja},
  journal={arXiv preprint arXiv:2510.02173},
  year={2025}
}

@article{seo2025verifying,
  title={Verifying the Verifiers: Unveiling Pitfalls and Potentials in Fact Verifiers},
  author={Seo, Wooseok and Han, Seungju and Jung, Jaehun and Newman, Benjamin and Lim, Seungwon and Lee, Seungbeen and Lu, Ximing and Choi, Yejin and Yu, Youngjae},
  journal={arXiv preprint arXiv:2506.13342},
  year={2025}
}

@inproceedings{song2024raghat,
  title={RAG-HAT: A hallucination-aware tuning pipeline for LLM in retrieval-augmented generation},
  author={Song, Juntong and Wang, Xingguang and Zhu, Juno and Wu, Yuanhao and Cheng, Xuxin and Zhong, Randy and Niu, Cheng},
  booktitle={Proceedings of the 2024 Conference on Empirical Methods in Natural Language Processing: Industry Track},
  pages={1548--1558},
  year={2024}
}

@article{chen2025retaining,
  title={Retaining by doing: The role of on-policy data in mitigating forgetting},
  author={Chen, Howard and Razin, Noam and Narasimhan, Karthik and Chen, Danqi},
  journal={arXiv preprint arXiv:2510.18874},
  year={2025}
}

@article{schulman2017ppo,
  title={Proximal policy optimization algorithms},
  author={Schulman, John and Wolski, Filip and Dhariwal, Prafulla and Radford, Alec and Klimov, Oleg},
  journal={arXiv preprint arXiv:1707.06347},
  year={2017}
}

@article{shao2024grpo,
  title={Deepseekmath: Pushing the limits of mathematical reasoning in open language models},
  author={Shao, Zhihong and Wang, Peiyi and Zhu, Qihao and Xu, Runxin and Song, Junxiao and Bi, Xiao and Zhang, Haowei and Zhang, Mingchuan and Li, YK and Wu, Yang and others},
  journal={arXiv preprint arXiv:2402.03300},
  year={2024}
}

@article{guo2025deepseekr1,
  title={Deepseek-r1: Incentivizing reasoning capability in llms via reinforcement learning},
  author={Guo, Daya and Yang, Dejian and Zhang, Haowei and Song, Junxiao and Zhang, Ruoyu and Xu, Runxin and Zhu, Qihao and Ma, Shirong and Wang, Peiyi and Bi, Xiao and others},
  journal={arXiv preprint arXiv:2501.12948},
  year={2025}
}

@article{zeng2025simplerl,
  title={Simplerl-zoo: Investigating and taming zero reinforcement learning for open base models in the wild},
  author={Zeng, Weihao and Huang, Yuzhen and Liu, Qian and Liu, Wei and He, Keqing and Ma, Zejun and He, Junxian},
  journal={arXiv preprint arXiv:2503.18892},
  year={2025}
}

@article{yu2025dapo,
  title={Dapo: An open-source llm reinforcement learning system at scale},
  author={Yu, Qiying and Zhang, Zheng and Zhu, Ruofei and Yuan, Yufeng and Zuo, Xiaochen and Yue, Yu and Dai, Weinan and Fan, Tiantian and Liu, Gaohong and Liu, Lingjun and others},
  journal={arXiv preprint arXiv:2503.14476},
  year={2025}
}

@article{zhao2025absolutezero,
  title={Absolute zero: Reinforced self-play reasoning with zero data},
  author={Zhao, Andrew and Wu, Yiran and Yue, Yang and Wu, Tong and Xu, Quentin and Lin, Matthieu and Wang, Shenzhi and Wu, Qingyun and Zheng, Zilong and Huang, Gao},
  journal={arXiv preprint arXiv:2505.03335},
  year={2025}
}

@article{liu2025deepseekV3,
  title={Deepseek-v3. 2: Pushing the frontier of open large language models},
  author={Liu, Aixin and Mei, Aoxue and Lin, Bangcai and Xue, Bing and Wang, Bingxuan and Xu, Bingzheng and Wu, Bochao and Zhang, Bowei and Lin, Chaofan and Dong, Chen and others},
  journal={arXiv preprint arXiv:2512.02556},
  year={2025}
}

@article{hurst2024gpt4o,
  title={Gpt-4o system card},
  author={Hurst, Aaron and Lerer, Adam and Goucher, Adam P and Perelman, Adam and Ramesh, Aditya and Clark, Aidan and Ostrow, AJ and Welihinda, Akila and Hayes, Alan and Radford, Alec and others},
  journal={arXiv preprint arXiv:2410.21276},
  year={2024}
}

@article{yang2025qwen3,
  title={Qwen3 technical report},
  author={Yang, An and Li, Anfeng and Yang, Baosong and Zhang, Beichen and Hui, Binyuan and Zheng, Bo and Yu, Bowen and Gao, Chang and Huang, Chengen and Lv, Chenxu and others},
  journal={arXiv preprint arXiv:2505.09388},
  year={2025}
}

@article{schmidhuber2013powerplay,
  title={Powerplay: Training an increasingly general problem solver by continually searching for the simplest still unsolvable problem},
  author={Schmidhuber, J{\"u}rgen},
  journal={Frontiers in psychology},
  volume={4},
  pages={313},
  year={2013},
  publisher={Frontiers Media SA}
}

@article{schaul2024boundless,
  title={Boundless socratic learning with language games},
  author={Schaul, Tom},
  journal={arXiv preprint arXiv:2411.16905},
  year={2024}
}

@article{silver2017mastering,
  title={Mastering chess and shogi by self-play with a general reinforcement learning algorithm},
  author={Silver, David and Hubert, Thomas and Schrittwieser, Julian and Antonoglou, Ioannis and Lai, Matthew and Guez, Arthur and Lanctot, Marc and Sifre, Laurent and Kumaran, Dharshan and Graepel, Thore and others},
  journal={arXiv preprint arXiv:1712.01815},
  year={2017}
}

@article{silver2017zero,
  title={Mastering the game of go without human knowledge},
  author={Silver, David and Schrittwieser, Julian and Simonyan, Karen and Antonoglou, Ioannis and Huang, Aja and Guez, Arthur and Hubert, Thomas and Baker, Lucas and Lai, Matthew and Bolton, Adrian and others},
  journal={nature},
  volume={550},
  number={7676},
  pages={354--359},
  year={2017},
  publisher={Nature Publishing Group UK London}
}

@article{goodfellow2020gan,
  title={Generative adversarial networks},
  author={Goodfellow, Ian and Pouget-Abadie, Jean and Mirza, Mehdi and Xu, Bing and Warde-Farley, David and Ozair, Sherjil and Courville, Aaron and Bengio, Yoshua},
  journal={Communications of the ACM},
  volume={63},
  number={11},
  pages={139--144},
  year={2020},
  publisher={ACM New York, NY, USA}
}

@article{chen2025spc,
  title={Spc: Evolving self-play critic via adversarial games for llm reasoning},
  author={Chen, Jiaqi and Zhang, Bang and Ma, Ruotian and Wang, Peisong and Liang, Xiaodan and Tu, Zhaopeng and Li, Xiaolong and Wong, Kwan-Yee K},
  journal={arXiv preprint arXiv:2504.19162},
  year={2025}
}

@article{kuba2025language,
  title={Language self-play for data-free training},
  author={Kuba, Jakub Grudzien and Gu, Mengting and Ma, Qi and Tian, Yuandong and Mohan, Vijai and Chen, Jason},
  journal={arXiv preprint arXiv:2509.07414},
  year={2025}
}

@article{jacovi2025facts,
  title={The FACTS Grounding Leaderboard: Benchmarking LLMs' Ability to Ground Responses to Long-Form Input},
  author={Jacovi, Alon and Wang, Andrew and Alberti, Chris and Tao, Connie and Lipovetz, Jon and Olszewska, Kate and Haas, Lukas and Liu, Michelle and Keating, Nate and Bloniarz, Adam and others},
  journal={arXiv preprint arXiv:2501.03200},
  year={2025}
}

@inproceedings{dhuliawala2024chain,
  title={Chain-of-verification reduces hallucination in large language models},
  author={Dhuliawala, Shehzaad and Komeili, Mojtaba and Xu, Jing and Raileanu, Roberta and Li, Xian and Celikyilmaz, Asli and Weston, Jason},
  booktitle={Findings of the association for computational linguistics: ACL 2024},
  pages={3563--3578},
  year={2024}
}

@article{cao2023autohall,
  title={Autohall: Automated hallucination dataset generation for large language models},
  author={Cao, Zouying and Yang, Yifei and Zhao, Hai},
  journal={arXiv preprint arXiv:2310.00259},
  year={2023}
}

@inproceedings{yang2018hotpotqa,
  title={HotpotQA: A dataset for diverse, explainable multi-hop question answering},
  author={Yang, Zhilin and Qi, Peng and Zhang, Saizheng and Bengio, Yoshua and Cohen, William and Salakhutdinov, Ruslan and Manning, Christopher D},
  booktitle={Proceedings of the 2018 conference on empirical methods in natural language processing},
  pages={2369--2380},
  year={2018}
}

@inproceedings{sheng2025verl,
  title={Hybridflow: A flexible and efficient rlhf framework},
  author={Sheng, Guangming and Zhang, Chi and Ye, Zilingfeng and Wu, Xibin and Zhang, Wang and Zhang, Ru and Peng, Yanghua and Lin, Haibin and Wu, Chuan},
  booktitle={Proceedings of the Twentieth European Conference on Computer Systems},
  pages={1279--1297},
  year={2025}
}

@inproceedings{zheng2024llamafactory,
  title={LlamaFactory: Unified Efficient Fine-Tuning of 100+ Language Models},
  author={Yaowei Zheng and Richong Zhang and Junhao Zhang and Yanhan Ye and Zheyan Luo and Zhangchi Feng and Yongqiang Ma},
  booktitle={Proceedings of the 62nd Annual Meeting of the Association for Computational Linguistics (Volume 3: System Demonstrations)},
  address={Bangkok, Thailand},
  publisher={Association for Computational Linguistics},
  year={2024},
  url={http://arxiv.org/abs/2403.13372}
}

@article{ragsurvey,
  title={A Survey on RAG with LLMs},
  author={Arslan, Muhammad and Ghanem, Hussam and Munawar, Saba and Cruz, Christophe},
  journal={Procedia computer science},
  volume={246},
  pages={3781--3790},
  year={2024},
  publisher={Elsevier}
}

@article{yang2025quantifying,
  title={Quantifying the robustness of retrieval-augmented language models against spurious features in grounding data},
  author={Yang, Shiping and Wu, Jie and Ding, Wenbiao and Wu, Ning and Liang, Shining and Gong, Ming and Zhang, Hengyuan and Zhang, Dongmei},
  journal={arXiv preprint arXiv:2503.05587},
  year={2025}
}

@article{halusurvey,
  title={A survey on hallucination in large language models: Principles, taxonomy, challenges, and open questions},
  author={Huang, Lei and Yu, Weijiang and Ma, Weitao and Zhong, Weihong and Feng, Zhangyin and Wang, Haotian and Chen, Qianglong and Peng, Weihua and Feng, Xiaocheng and Qin, Bing and others},
  journal={ACM Transactions on Information Systems},
  volume={43},
  number={2},
  pages={1--55},
  year={2025},
  publisher={ACM New York, NY}
}

@inproceedings{tan2024large,
  title={Large language models for data annotation and synthesis: A survey},
  author={Tan, Zhen and Li, Dawei and Wang, Song and Beigi, Alimohammad and Jiang, Bohan and Bhattacharjee, Amrita and Karami, Mansooreh and Li, Jundong and Cheng, Lu and Liu, Huan},
  booktitle={Proceedings of the 2024 Conference on Empirical Methods in Natural Language Processing},
  pages={930--957},
  year={2024}
}

@inproceedings{li2025generation,
  title={From generation to judgment: Opportunities and challenges of llm-as-a-judge},
  author={Li, Dawei and Jiang, Bohan and Huang, Liangjie and Beigi, Alimohammad and Zhao, Chengshuai and Tan, Zhen and Bhattacharjee, Amrita and Jiang, Yuxuan and Chen, Canyu and Wu, Tianhao and others},
  booktitle={Proceedings of the 2025 Conference on Empirical Methods in Natural Language Processing},
  pages={2757--2791},
  year={2025}
}

@article{zhang2025find,
  title={Find Your Optimal Teacher: Personalized Data Synthesis via Router-Guided Multi-Teacher Distillation},
  author={Zhang, Hengyuan and Yang, Shiping and Liang, Xiao and Shang, Chenming and Jiang, Yuxuan and Tao, Chaofan and Xiong, Jing and So, Hayden Kwok-Hay and Xie, Ruobing and Chang, Angel X and others},
  journal={arXiv preprint arXiv:2510.10925},
  year={2025}
}

@article{liang2026sws,
  title={Sws: Self-aware weakness-driven problem synthesis in reinforcement learning for llm reasoning},
  author={Liang, Xiao and Li, Zhong-Zhi and Gong, Yeyun and Wang, Yang and Zhang, Hengyuan and Wu, Ying Nian and Chen, Weizhu and others},
  journal={Advances in Neural Information Processing Systems},
  volume={38},
  pages={56801--56839},
  year={2026}
}

@inproceedings{zha-etal-2023-alignscore,
    title = "{A}lign{S}core: Evaluating Factual Consistency with A Unified Alignment Function",
    author = "Zha, Yuheng  and
      Yang, Yichi  and
      Li, Ruichen  and
      Hu, Zhiting",
    editor = "Rogers, Anna  and
      Boyd-Graber, Jordan  and
      Okazaki, Naoaki",
    booktitle = "Proceedings of the 61st Annual Meeting of the Association for Computational Linguistics (Volume 1: Long Papers)",
    month = jul,
    year = "2023",
    address = "Toronto, Canada",
    publisher = "Association for Computational Linguistics",
    url = "https://aclanthology.org/2023.acl-long.634/",
    doi = "10.18653/v1/2023.acl-long.634",
    pages = "11328--11348"
}

@article{ravi2024lynx,
  title={Lynx: An open source hallucination evaluation model},
  author={Ravi, Selvan Sunitha and Mielczarek, Bartosz and Kannappan, Anand and Kiela, Douwe and Qian, Rebecca},
  journal={arXiv preprint arXiv:2407.08488},
  year={2024}
}

@inproceedings{si2026faithlens,
  title={Faithlens: Detecting and explaining faithfulness hallucination},
  author={Si, Shuzheng and Wang, Qingyi and Zhao, Haozhe and Bai, Yuzhuo and Chen, Guanqiao and Luo, Kangyang and Chen, Gang and Qi, Fanchao and Zhang, Minjia and Chang, Baobao and others},
  booktitle={Findings of the Association for Computational Linguistics: ACL 2026},
  pages={14068--14099},
  year={2026}
}

@inproceedings{utama2022falsesum,
  title={Falsesum: Generating document-level NLI examples for recognizing factual inconsistency in summarization},
  author={Utama, Prasetya and Bambrick, Joshua and Moosavi, Nafise Sadat and Gurevych, Iryna},
  booktitle={Proceedings of the 2022 Conference of the North American Chapter of the Association for Computational Linguistics: Human Language Technologies},
  pages={2763--2776},
  year={2022}
}

@inproceedings{belyi-etal-2025-luna,
    title = "{L}una: A Lightweight Evaluation Model to Catch Language Model Hallucinations with High Accuracy and Low Cost",
    author = "Belyi, Masha  and
      Friel, Robert  and
      Shao, Shuai  and
      Sanyal, Atindriyo",
    editor = "Rambow, Owen  and
      Wanner, Leo  and
      Apidianaki, Marianna  and
      Al-Khalifa, Hend  and
      Eugenio, Barbara Di  and
      Schockaert, Steven  and
      Darwish, Kareem  and
      Agarwal, Apoorv",
    booktitle = "Proceedings of the 31st International Conference on Computational Linguistics: Industry Track",
    month = jan,
    year = "2025",
    address = "Abu Dhabi, UAE",
    publisher = "Association for Computational Linguistics",
    url = "https://aclanthology.org/2025.coling-industry.34/",
    pages = "398--409"
}

@article{kovacs2025lettucedetect,
  title={Lettucedetect: A hallucination detection framework for rag applications},
  author={Kov{\'a}cs, {\'A}d{\'a}m and Recski, G{\'a}bor},
  journal={arXiv preprint arXiv:2502.17125},
  year={2025}
}

@inproceedings{huang2026r,
  title={R-zero: Self-evolving reasoning llm from zero data},
  author={Huang, Chengsong and Yu, Wenhao and Wang, Xiaoyang and Zhang, Hongming and Li, Zongxia and Li, Ruosen and Huang, Jiaxin and Mi, Haitao and Yu, Dong},
  booktitle={International Conference on Learning Representations},
  volume={2026},
  pages={130770--130790},
  year={2026}
}

@inproceedings{manakul2023selfcheckgpt,
  title={Selfcheckgpt: Zero-resource black-box hallucination detection for generative large language models},
  author={Manakul, Potsawee and Liusie, Adian and Gales, Mark},
  booktitle={Proceedings of the 2023 conference on empirical methods in natural language processing},
  pages={9004--9017},
  year={2023}
}

@article{cheng2024self,
  title={Self-playing adversarial language game enhances llm reasoning},
  author={Cheng, Pengyu and Dai, Yong and Hu, Tianhao and Xu, Han and Zhang, Zhisong and Han, Lei and Du, Nan and Li, Xiaolong},
  journal={Advances in Neural Information Processing Systems},
  volume={37},
  pages={126515--126543},
  year={2024}
}

@article{achiam2023gpt,
  title={Gpt-4 technical report},
  author={Achiam, Josh and Adler, Steven and Agarwal, Sandhini and Ahmad, Lama and Akkaya, Ilge and Aleman, Florencia Leoni and Almeida, Diogo and Altenschmidt, Janko and Altman, Sam and Anadkat, Shyamal and others},
  journal={arXiv preprint arXiv:2303.08774},
  year={2023}
}

@inproceedings{kandpal2023large,
  title={Large language models struggle to learn long-tail knowledge},
  author={Kandpal, Nikhil and Deng, Haikang and Roberts, Adam and Wallace, Eric and Raffel, Colin},
  booktitle={International conference on machine learning},
  pages={15696--15707},
  year={2023},
  organization={PMLR}
}

@inproceedings{vu2024freshllms,
  title={Freshllms: Refreshing large language models with search engine augmentation},
  author={Vu, Tu and Iyyer, Mohit and Wang, Xuezhi and Constant, Noah and Wei, Jerry and Wei, Jason and Tar, Chris and Sung, Yun-Hsuan and Zhou, Denny and Le, Quoc and others},
  booktitle={Findings of the Association for Computational Linguistics: ACL 2024},
  pages={13697--13720},
  year={2024}
}

@inproceedings{knowledge-conflicts,
    title = "Knowledge Conflicts for {LLM}s: A Survey",
    author = "Xu, Rongwu  and
      Qi, Zehan  and
      Guo, Zhijiang  and
      Wang, Cunxiang  and
      Wang, Hongru  and
      Zhang, Yue  and
      Xu, Wei",
    editor = "Al-Onaizan, Yaser  and
      Bansal, Mohit  and
      Chen, Yun-Nung",
    booktitle = "Proceedings of the 2024 Conference on Empirical Methods in Natural Language Processing",
    month = nov,
    year = "2024",
    address = "Miami, Florida, USA",
    publisher = "Association for Computational Linguistics",
    url = "https://aclanthology.org/2024.emnlp-main.486/",
    doi = "10.18653/v1/2024.emnlp-main.486",
    pages = "8541--8565"
}

@article{yang2026modularized,
  title={Modularized Reinforcement Learning on LLMs: From MDP Creation to Exploration and Learning},
  author={Yang, Zhao and Jiang, Yuxuan and Chen, Ting-Chih and Yang, Lincen and Wong, Annie and Gao, Chao and Kooi, Jacob E and Li, Zhong and Shi, Jiayang and Qiu, Kevin and others},
  journal={arXiv preprint arXiv:2606.21943},
  year={2026}
}
\bibliographystyle{colm2026_conference}

\clearpage
\appendix

\section{Implementation Details}
\label{sec:implementation_details}

All experiments are conducted on 8 NVIDIA A100 GPUs. 
We implement SFT training using \textbf{LlamaFactory}~\citep{zheng2024llamafactory} and RL training using the \textbf{verl} framework~\citep{sheng2025verl}. 
Each training stage for both the detector and generator uses 10k samples.
The detailed hyperparameters for both training stages are summarized in Table~\ref{tab:hyperparams}.

\begin{table}[h]
\centering
\begin{tabular}{lcc}
\hline
\textbf{Hyperparameter} & \textbf{SFT} & \textbf{RL (GRPO)} \\
\hline
Learning rate & 1e-5 & 1e-6 \\
LR scheduler & cosine & cosine \\
Warmup & 0.1 (ratio) & 5 (steps) \\
Train batch size & 32 & 256 \\
Epochs & 2 & 1 \\
Mini-batch size & --- & 64 \\
Rollout samples ($K$) & --- & 8 \\
\hline
\end{tabular}
\caption{Training hyperparameters for SFT and RL stages.}
\label{tab:hyperparams}
\end{table}

\section{Prompt Templates}
\label{sec:prompt_templates}

\begin{figure*}[htbp]
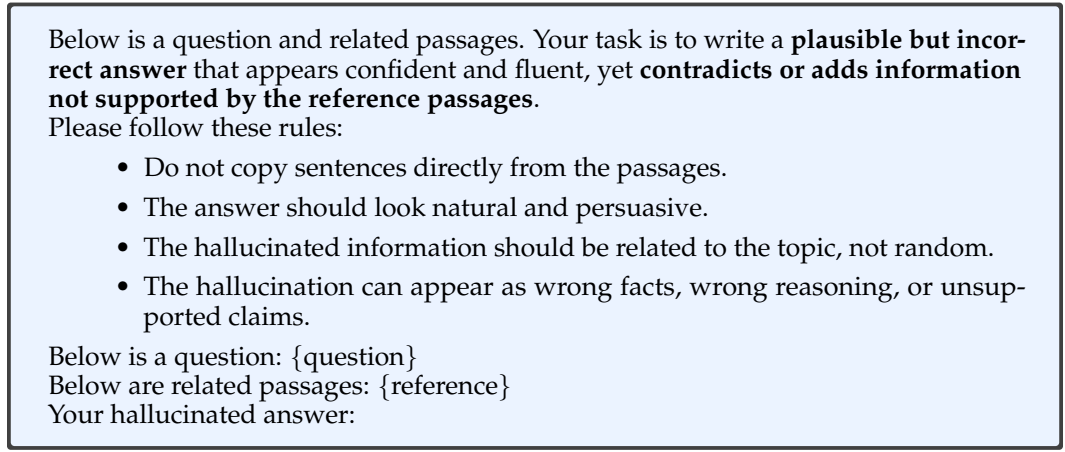

\begin{tcolorbox}[
    colback=blue-back,
    boxrule=0.5mm,
    arc=0.5mm,
    outer arc=0.5mm]
Below is a question and related passages.
Your task is to write a \textbf{plausible but incorrect answer} that appears confident and fluent,
yet \textbf{contradicts or adds information not supported by the reference passages}.

Please follow these rules:
\begin{itemize}
    \item Do not copy sentences directly from the passages.
    \item The answer should look natural and persuasive.
    \item The hallucinated information should be related to the topic, not random.
    \item The hallucination can appear as wrong facts, wrong reasoning, or unsupported claims.
\end{itemize}

Below is a question: \{question\}

Below are related passages: \{reference\}

Your hallucinated answer:
\end{tcolorbox}
\caption{Prompt template for hallucination generator.}
\label{fig:generator_template}
\end{figure*}

\begin{figure*}[htbp]
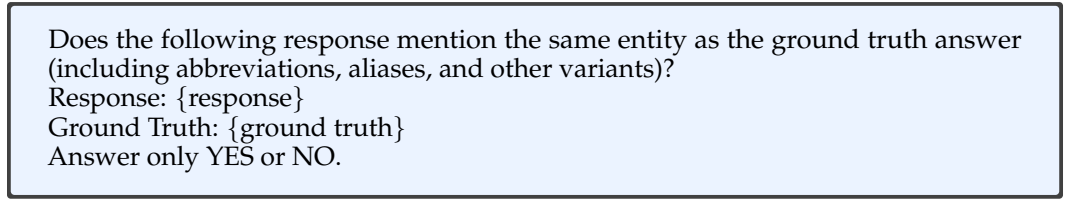

\begin{tcolorbox}[
    colback=blue-back,
    boxrule=0.5mm,
    arc=0.5mm,
    outer arc=0.5mm]
Does the following response mention the same entity as the ground truth answer (including abbreviations, aliases, and other variants)?

Response: \{response\}

Ground Truth: \{ground truth\}

Answer only YES or NO.
\end{tcolorbox}
\caption{Prompt template for model-based answer alias detection, used as part of the reward gating criteria.}
\label{fig:alias_detection_prompt}
\end{figure*}

\begin{figure*}[htbp]
\begin{tcolorbox}[
    colback=blue-back,
    boxrule=0.5mm,
    arc=0.5mm,
    outer arc=0.5mm]
You are given a question and a model-generated answer.

Your task is to determine whether the answer is an \textbf{EVASIVE ANSWER}.

\textbf{Definition:} An evasive answer is one that:
\begin{itemize}
    \item Does NOT directly answer the key attribute(s) explicitly asked in the question, AND
    \item Avoids providing the required information by giving vague descriptions, generic restatements, or partial information.
\end{itemize}

An evasive answer is NOT the same as a hallucinated answer.
\begin{itemize}
    \item Hallucination introduces incorrect facts.
    \item Evasion avoids answering the question.
\end{itemize}

You should output:
\begin{itemize}
    \item ``EVASIVE'' if the answer fails to provide the required information.
    \item ``NOT EVASIVE'' if the answer directly answers the question.
\end{itemize}

Focus strictly on whether the \textbf{question is answered}, not on correctness of background details.

\tcblower

\textbf{Question:} Claude-Auguste Lamy discovered the element thallium independently from this English chemist who died in what year?

\textbf{Answer:} Claude-Auguste Lamy discovered the element thallium independently from William Crookes, who was a prominent English chemist active in the mid-19th century.

\textbf{Explanation:} The question explicitly asks for the year of death. The answer identifies the chemist but does not provide the requested year. 

\textbf{Label: EVASIVE}

\medskip
\textbf{Question:} Claude-Auguste Lamy discovered the element thallium independently from this English chemist who died in what year?

\textbf{Answer:} Claude-Auguste Lamy discovered thallium independently from William Crookes, who died in 1905.

\textbf{Explanation:} The answer directly provides the requested information (a year of death), but the year is incorrect. This is a hallucinated answer, not an evasive one. 

\textbf{Label: NOT EVASIVE}

\medskip
\textbf{Question:} \{question\}

\textbf{Answer:} \{response\}

\textbf{Explanation:}

\textbf{Label:}
\end{tcolorbox}
\caption{Prompt template for model-based evasive answer detection, used as part of the trivial answer penalty.}
\label{fig:evasive_detection_prompt}
\end{figure*}

\subsection{Detector Prompt Template}

\begin{figure*}[htbp]
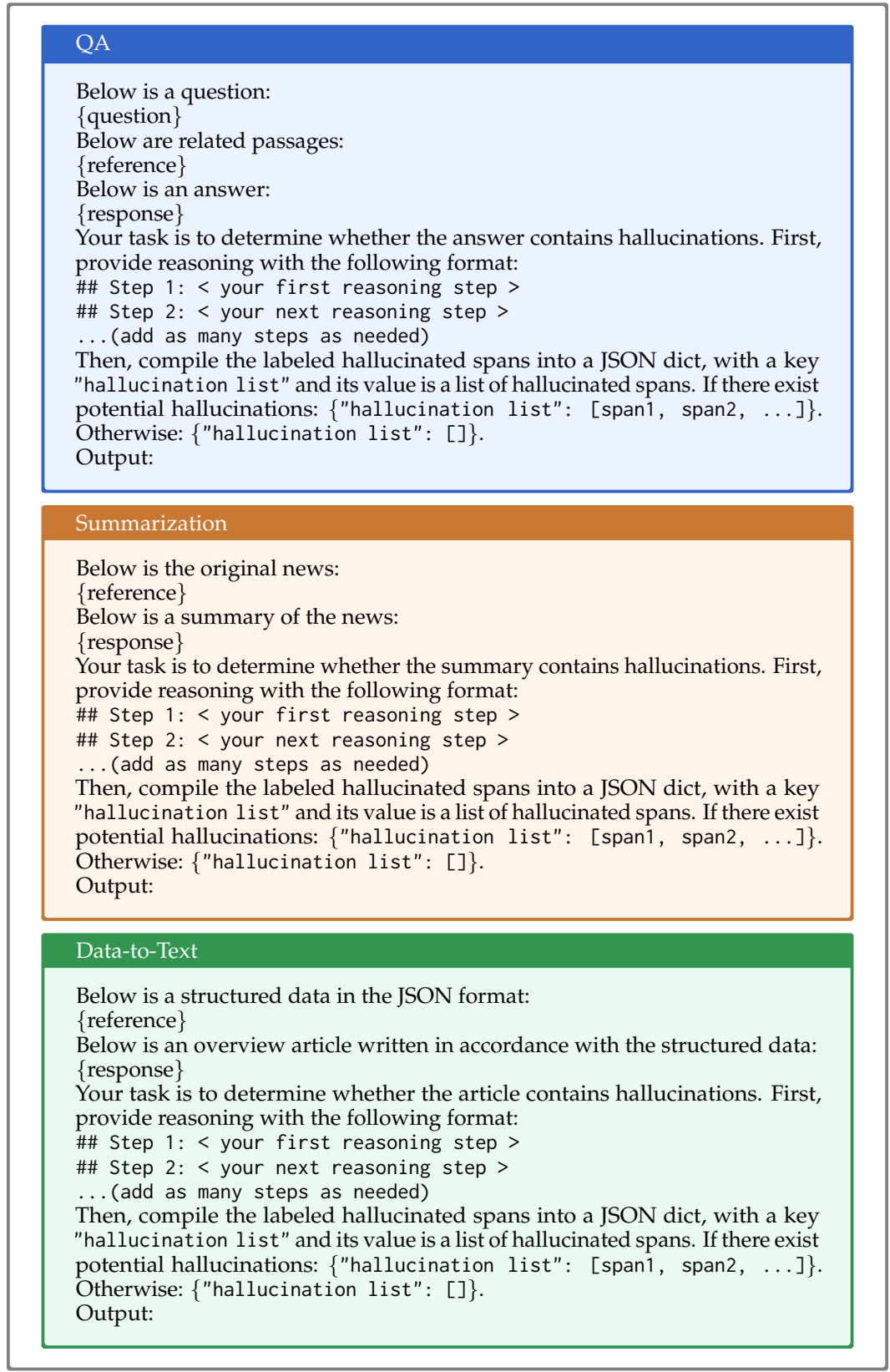

\begin{tcolorbox}[
    colback=white,
    colframe=black!50,
    boxrule=0.5mm,
    arc=0.5mm,
    outer arc=0.5mm,
]
\begin{tcolorbox}[
    boxrule=0.5mm,
    arc=0.5mm,
    outer arc=0.5mm,
    colback=blue-back,
    colframe=blue-title,
    title=QA
]
Below is a question:\\
\{question\}

Below are related passages:\\
\{reference\}

Below is an answer:\\
\{response\}

Your task is to determine whether the answer contains hallucinations.
First, provide reasoning with the following format:\\
\texttt{\#\# Step 1: < your first reasoning step >}\\
\texttt{\#\# Step 2: < your next reasoning step >}\\
\texttt{...(add as many steps as needed)}

Then, compile the labeled hallucinated spans into a JSON dict, with a key \texttt{"hallucination list"} and its value is a list of hallucinated spans. If there exist potential hallucinations: \texttt{\{"hallucination list": [span1, span2, ...]\}}. Otherwise: \texttt{\{"hallucination list": []\}}.

Output:
\end{tcolorbox}

\begin{tcolorbox}[
    boxrule=0.5mm,
    arc=0.5mm,
    outer arc=0.5mm,
    colback=orange-back,
    colframe=orange-title,
    title=Summarization
]
Below is the original news:\\
\{reference\}

Below is a summary of the news:\\
\{response\}

Your task is to determine whether the summary contains hallucinations.
First, provide reasoning with the following format:\\
\texttt{\#\# Step 1: < your first reasoning step >}\\
\texttt{\#\# Step 2: < your next reasoning step >}\\
\texttt{...(add as many steps as needed)}

Then, compile the labeled hallucinated spans into a JSON dict, with a key \texttt{"hallucination list"} and its value is a list of hallucinated spans. If there exist potential hallucinations: \texttt{\{"hallucination list": [span1, span2, ...]\}}. Otherwise: \texttt{\{"hallucination list": []\}}.

Output:
\end{tcolorbox}

\begin{tcolorbox}[
    boxrule=0.5mm,
    arc=0.5mm,
    outer arc=0.5mm,
    colback=green-back,
    colframe=green-title,
    title=Data-to-Text
]
Below is a structured data in the JSON format:\\
\{reference\}

Below is an overview article written in accordance with the structured data:\\
\{response\}

Your task is to determine whether the article contains hallucinations.
First, provide reasoning with the following format:\\
\texttt{\#\# Step 1: < your first reasoning step >}\\
\texttt{\#\# Step 2: < your next reasoning step >}\\
\texttt{...(add as many steps as needed)}

Then, compile the labeled hallucinated spans into a JSON dict, with a key \texttt{"hallucination list"} and its value is a list of hallucinated spans. If there exist potential hallucinations: \texttt{\{"hallucination list": [span1, span2, ...]\}}. Otherwise: \texttt{\{"hallucination list": []\}}.

Output:
\end{tcolorbox}
\end{tcolorbox}
\caption{Prompt templates used for the detector w/ CoT across three task types: QA, Summarization, and Data-to-Text.}
\label{fig:detector_template}
\end{figure*}

\begin{table}[h]
\centering
\begin{tabular}{lccc}
\hline
\textbf{Variant} & \textbf{Hallucination Rate} & \textbf{Avg. Reward} & \textbf{Trainable Samples} \\
\hline
HSP (Full)                  & 90\% & 0.288 & 91 / 500  \\
HSP w/o gating              &  0\% & 1.000 & 500 / 500 \\
HSP w/o trivial penalty     & 70\% & 0.333 & 277 / 500 \\
HSP w/o both                &  0\% & 1.000 & 500 / 500 \\
\hline
\end{tabular}
\caption{Ablation on reward hacking mitigation mechanisms. \textit{Hallucination Rate} measures the fraction of generated responses that are genuinely hallucinated. \textit{Trainable Samples} denotes the number of samples with positive reward retained for detector training.}
\label{tab:ablation_reward_hacking}
\end{table}

\end{document}